\documentclass[runningheads]{llncs}
\usepackage[T1]{fontenc}
\usepackage{graphicx,verbatim}
\usepackage{amsmath}
\usepackage{amssymb}
\usepackage{graphicx} 
\usepackage{booktabs}  
\usepackage{multirow}   
\usepackage{siunitx}  
\usepackage{color}
\usepackage{hyperref}

\usepackage{url}
\begin{document}
\title{TAVR-VLM: Risk-Conditioned Causal Grounding for Hallucination-Resistant Report Generation}
\titlerunning{TAVR-VLM}

\author{Zhixiang Lu\inst{1,4} \and
Xiwei Liu\inst{2} \and
Sifan Song\inst{1} \and
Changkai Ji\inst{3} \and
Anh Nguyen\inst{4} \and
Jionglong Su\inst{1} \and
Imran Razzak\inst{2} \and
Jinfeng Wang\inst{1,5}}
\authorrunning{Z. Lu et al.}
%
\institute{Xi'an Jiaotong-Liverpool University\\
 \and
Mohamed bin Zayed University of Artificial Intelligence\\
 \and
 Shanghai Jiao Tong University\\
 \and
University of Liverpool\\
 \and
 Kunming University of Science and Technology\\
\email{zhixiang@liverpool.ac.uk}}

\maketitle              
\begin{abstract}
Transcatheter Aortic Valve Replacement (TAVR) planning requires meticulous multimodal reasoning. However, adapting Multimodal Large Language Models (MLLMs) to this high-stakes domain is severely impeded by diagnostic hallucinations, where generated text lacks anatomical grounding. To address this, TAVR-VLM is introduced: a novel framework featuring Risk-Conditioned Causal Grounding Attention (R-CGA) that instantiates a model-internal ``Risk $\rightarrow$ Region $\rightarrow$ Word'' structural grounding pathway. R-CGA compresses multimodal inputs into a causal risk bottleneck, purifying dense visual features into a global risk mask. During autoregressive generation, a support-projected causal consistency objective constrains token-level grounding within the risk-defined support mask. Evaluated on $\text{M}^3\text{TAVR}$, a comprehensive 1,482-patient cohort, TAVR-VLM establishes a new state-of-the-art. It achieves an AUROC of 0.896, boosts CIDEr to 0.936, and drastically reduces the hallucination rate to 8.1\%, thereby improving interpretability for evidence-based surgical AI.
\end{abstract}


%
%
%
\section{Introduction}

Transcatheter Aortic Valve Replacement (TAVR) has established itself as the standard-of-care for severe aortic stenosis across the full spectrum of surgical risk \cite{leon2010partner,mack2019partner3}. Despite its clinical success, procedural outcomes remain highly sensitive to the precision of preoperative planning. Subtle inaccuracies in the assessment of 3D CT annular geometry or 2D echocardiographic hemodynamics can precipitate catastrophic complications, including annular rupture, coronary obstruction, or significant paravalvular leak \cite{varc32021,lilly2020conduction}. To ensure safety and reproducibility, the Valve Academic Research Consortium (VARC-3) emphasizes the necessity of evidence-grounded evaluation \cite{varc32021}. Consequently, an ideal AI-driven TAVR system must transcend simple risk scoring to provide structured clinical reports and actionable recommendations explicitly grounded in multimodal imaging evidence.

The rapid evolution of multimodal large language models (MLLMs) has introduced the potential for generalist medical assistants capable of joint image-text interpretation \cite{li2023llavamed}. While radiology report generation has progressed significantly using Transformer-based architectures on large-scale datasets \cite{johnson2019mimiccxr,lu2026skinclipvlconsistencyawarevisionlanguagelearning}, adapting these models to TAVR-grade clinical reasoning reveals a critical technical bottleneck. TAVR planning requires a rigorous coupling between global procedural risk states and localized anatomical findings. Contemporary medical MLLMs frequently exhibit diagnostic hallucinations, where the model generates plausible but non-existent clinical findings or recommendations unsupported by the underlying imagery \cite{liu2024lvhallsurvey,lu2026dialecticmedmitigatingdiagnostichallucinations}. In the context of structural heart interventions, such hallucinations are unacceptable as they directly compromise surgical decision-making. We posit that hallucinations persist in medical MLLMs due to the absence of explicit causal constraints \cite{causalsamllm,xue2026semantictopologicalgraphreasoninglanguageguided}. Traditional models typically generate text by attending to dense visual tokens without ensuring that each reported clinical entity is anchored to risk-relevant evidence \cite{lu2026dialecticmedmitigatingdiagnostichallucinations,lu2026skinclipvlconsistencyawarevisionlanguagelearning}. In professional TAVR workflows, clinicians utilize a top-down reasoning paradigm: they first establish a global risk hypothesis, then systematically interrogate specific anatomical regions to seek supporting evidence, and finally synthesize these insights into a structured plan \cite{hierrisk,lu2026dialecticmedmitigatingdiagnostichallucinations}. This observation motivates our core design principle: Risk should serve as a causal bottleneck that governs the visual perception and linguistic output of the generator.

To operationalize this principle, we introduce \textbf{Risk-Conditioned Causal Grounding Attention (R-CGA)}, a top-down closed-loop module that integrates risk stratification with multi-granularity grounding for TAVR planning. R-CGA first predicts a global risk distribution and compresses it into a learnable, causal bottleneck. This bottleneck is then utilized to purify dense visual features into a global risk mask, ensuring the model encourages the model to focus on risk-relevant evidence relevant to the predicted risk state. Finally, we introduce a causal consistency supervision mechanism with a stop-gradient regularizer. This ensures that the token-level grounding maps for all generated clinical entities are spatially contained within the global risk-evidence region, thereby transforming risk prediction into a structural prior that mitigates hallucinations. Our main contributions are summarized as follows:

\begin{itemize}
    \item We formalize TAVR assessment as a risk-conditioned structured generation task, identifying hallucination as a failure of the top-down causal link between global risk states and local anatomical evidence.
    \item We propose a three-stage architecture comprising a causal risk bottleneck, risk-to-region evidence selection, and region-to-word token grounding to enforce a strict ``Risk $\rightarrow$ Region $\rightarrow$ Word'' reasoning hierarchy.
    \item We conduct comprehensive experiments on the $\text{M}^3\text{TAVR}$, including risk prediction, report generation, hallucination evaluation, spatial grounding, ablation analysis and qualitative grounding analysis.
\end{itemize}

\section{Methodology}
\label{sec:method}

\subsection{Problem Formulation and Causal Framework}
\label{sec:setup}
\begin{figure}[t]
\centering
\includegraphics[width=\textwidth]{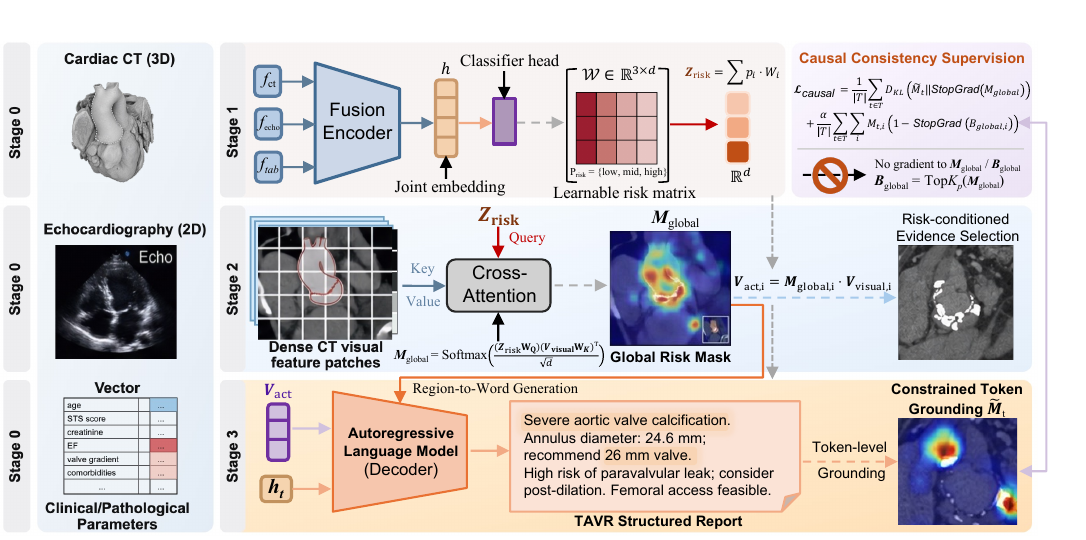}
\caption{\textbf{Overview of the TAVR-VLM architecture.} R-CGA implements a model-internal
``Risk $\rightarrow$ Region $\rightarrow$ Word'' structural grounding pathway.
\textbf{Stage 1} encodes 3D CT, echocardiography clips, and clinical parameters into a
joint embedding, predicts a risk distribution $P_{\mathrm{risk}}$, and constructs
a causal risk bottleneck $Z_{\mathrm{risk}}=P_{\mathrm{risk}}^{\top}W$.
\textbf{Stage 2} uses $Z_{\mathrm{risk}}$ as the query and CT visual tokens as key/value
features to obtain a global risk mask $M_{\mathrm{global}}$, from which a
risk-defined support mask $B_{\mathrm{global}}=\mathrm{TopK}{\rho}
(M{\mathrm{global}})$ is derived. \textbf{Stage 3} generates the structured report from
risk-activated visual features $V_{\mathrm{act}}$ while projecting raw token
attention $M_t$ into $B_{\mathrm{global}}$ to obtain the constrained token mask
$\widetilde{M}_t$. The causal consistency loss aligns $\widetilde{M}_t$ with the
global risk mask and penalizes attention outside the risk-defined support.}
\label{fig:framework}
\end{figure}

We study TAVR pre-operative decision support as a risk-conditioned multimodal generation problem. Given multimodal inputs $\mathcal{X} = \{X_{\text{CT}}, X_{\text{Echo}}, X_{\text{tab}}\}$ (3D CT, 2D echocardiography, and clinical tables), our objective is to learn a joint distribution $P(Y, r \mid \mathcal{X})$. Here, $r \in \{l, m, h\}$ denotes the discrete procedural risk level, and $Y = (y_1, \dots, y_T)$ represents the autoregressive structured report. To mitigate diagnostic hallucinations, we depart from standard multi-task learning by introducing a strict top-down causal constraint: $\text{Risk} \rightarrow \text{Region} \rightarrow \text{Word}$. We operationalize this via the \textbf{Risk-Conditioned Causal Grounding Attention (R-CGA)} module, which enforces that every clinically salient entity in $Y$ is intrinsically grounded in risk-relevant anatomical evidence (Fig.~\ref{fig:framework}).

\subsection{Risk-Conditioned Causal Grounding Attention (R-CGA)}
\label{sec:rcga}
R-CGA implements the causal hierarchy through three interconnected stages, acting as an information bottleneck that filters irrelevant visual noise prior to text generation.

\paragraph{Stage 1: Causal Risk Bottleneck.}
Modality-specific encoders map inputs into a unified latent space, extracting dense visual tokens $V \in \mathbb{R}^{N \times d}$. A lightweight fusion network computes a joint patient embedding to predict the risk distribution $P_{\text{risk}} = [p_l, p_m, p_h]^\top$. Instead of passing raw embeddings to the generator, we introduce a learnable risk matrix $W \in \mathbb{R}^{3 \times d}$. The \textbf{Causal Risk Bottleneck} is computed as the expected prototype under $P_{\text{risk}}$:
\begin{equation}
    Z_{\text{risk}} = P_{\text{risk}}^\top W \in \mathbb{R}^d.
\end{equation}
$Z_{\text{risk}}$ serves as a distribution-aware macro ``commander'' state, encapsulating the minimal sufficient clinical evidence required to justify the predicted risk profile.

\paragraph{Stage 2: Risk-to-Region Evidence Selection.}
Let $V_{\mathrm{visual}}\in\mathbb{R}^{N\times d}$ denote the dense CT visual
tokens used for spatial grounding. The risk bottleneck $Z_{\mathrm{risk}}$ acts
as the query, while $V_{\mathrm{visual}}$ serves as key/value features. The
global risk mask is obtained by cross-attention \cite{gheini-etal-2021-cross,deepgbtb2026}:
\begin{equation}
M_{\mathrm{global}}
=
\mathrm{Softmax}
\left(
\frac{
(Z_{\mathrm{risk}}W_Q)(V_{\mathrm{visual}}W_K)^{\top}
}{\sqrt{d_k}}
\right)
\in \Delta^{N-1}.
\end{equation}
The risk-activated visual features are then computed by token-wise gating:
\begin{equation}
V_{\mathrm{act},i}
=
M_{\mathrm{global},i}\cdot V_{\mathrm{visual},i},
\qquad i=1,\ldots,N.
\end{equation}
This separates the soft global risk mask used for visual activation from the support mask used for strict token containment.

\paragraph{Stage 3: Region-to-Word Token Grounding.}
During autoregressive generation, the decoder hidden state $h_t$ computes a raw token-level attention map over the original CT visual tokens:
\begin{equation}
M_t
=
\mathrm{Softmax}
\left(
\frac{
(h_tW_Q^{(t)})(V_{\mathrm{visual}}W_K^{(t)})^{\top}
}{\sqrt{d_k}}
\right)
\in \Delta^{N-1}.
\end{equation}
Let $\mathcal{T}$ denote the set of token indices corresponding to clinically
salient entities, such as ``calcification'', ``LVOT''. For each $t\in\mathcal{T}$, $M_t$ represents the raw visual evidence used by the decoder before causal support projection.

\subsection{Support-Projected Causal Consistency and Optimization}

The global risk mask $M_{\text{global}}$ identifies the anatomical region that is relevant to the predicted risk state. To enforce that clinically salient tokens are grounded within this region, we derive a risk-defined support mask:
\begin{equation}
    B_{\text{global}} = \text{TopK}_{\rho}(M_{\text{global}}),
\end{equation}
where $\text{TopK}_{\rho}(\cdot)$ retains the top $\rho$ proportion of visual tokens and returns a binary support mask. We then project the raw token grounding map $M_t$ into this risk-defined support:
\begin{equation}
    \widetilde{M}_t = \frac{M_t \odot \text{StopGrad}(B_{\text{global}})}{\sum_i M_{t,i}\text{StopGrad}(B_{\text{global},i})+\epsilon}.
\end{equation}
By construction,
\begin{equation}
    \text{supp}(\widetilde{M}_t) \subseteq \text{supp}(B_{\text{global}}), \quad \forall t \in \mathcal{T}.
\end{equation}
This support projection provides the formal containment property that the raw KL alignment alone cannot guarantee. The final \textbf{Support-Projected Causal Consistency Loss} $\mathcal{L}_{\text{causal}}$ contains two terms:
\begin{equation}
    \begin{aligned}
    \mathcal{L}_{\text{causal}} &= \frac{1}{|\mathcal{T}|} \sum_{t\in\mathcal{T}} \mathbb{D}_{\text{KL}}\big(\widetilde{M}_t \parallel \text{StopGrad}(M_{\text{global}})\big) \\
    &\quad + \frac{\alpha}{|\mathcal{T}|} \sum_{t\in\mathcal{T}}\sum_i M_{t,i} \big(1-\text{StopGrad}(B_{\text{global},i})\big).
    \end{aligned}
\end{equation}
The first term aligns the constrained token grounding distribution with the global risk mask. The second term penalizes raw token attention that leaks outside the risk-defined support. The $\text{StopGrad}$ operator is applied only to the causal consistency branch, so that token grounding is optimized to obey the risk-defined region without allowing the token loss to expand or diffuse the global risk support. The end-to-end multi-task objective is jointly optimized:
\begin{equation}
    \mathcal{L} = \mathcal{L}_{\text{risk}} + \mathcal{L}_{\text{LM}} + \lambda_{\text{rec}} \mathcal{L}_{\text{rec}} + \lambda_{\text{causal}} \mathcal{L}_{\text{causal}},
\end{equation}
where $\mathcal{L}_{\text{risk}}$ is the cross-entropy for risk prediction, $\mathcal{L}_{\text{LM}}$ is the autoregressive negative log-likelihood over $V_{\text{act}}$, and $\mathcal{L}_{\text{rec}}$ optimizes surgical recommendations. This design fundamentally transforms risk prediction from an isolated classification head into a structural prior that governs generation.
\section{Experiments}
\label{sec:experiments}

\subsection{Dataset}
\label{sec:dataset}

To rigorously evaluate our framework, we curate \textbf{$\text{M}^3\text{TAVR}$} (Multimodal, Multi-granularity, Multi-task cohort for TAVR), the largest structural heart disease benchmark to date, comprising 1,482 patients. Each patient record is strictly paired with three modalities: (i) 3D cardiac CT volumes, (ii) 2D echocardiography clips, and (iii) a comprehensive vector of clinical and pathological biomarkers $X_{\text{tab}}$. Furthermore, each case is annotated with a ground-truth procedural risk level $r \in \{l, m, h\}$ and a clinician-authored structured report $Y$ containing diagnostic findings and surgical recommendations (e.g., valve sizing and access routes). To prevent data leakage and ensure robust evaluation, we employ a strict patient-level stratified split: 70\% for training, 10\% for validation, and 20\% for testing. For a randomly sampled subset of the test set ($N=200$), we additionally obtained expert-annotated region-of-interests (ROIs) for TAVR-critical structures to enable quantitative grounding evaluation.

\begin{table*}[t]
\centering
\caption{Comprehensive benchmarking on the \textbf{M$^3$TAVR} test set. Models are grouped into four paradigms. Best results are in \textbf{bold}, and second-best are \underline{underlined}.}
\label{tab:main}
\resizebox{\textwidth}{!}{
\begin{tabular}{l|cc|ccc|cc}
\toprule
\multirow{2}{*}{\textbf{Model}} & \multicolumn{2}{c|}{\textbf{Risk Prediction}} & \multicolumn{3}{c|}{\textbf{Report Generation}} & \multicolumn{2}{c}{\textbf{Safety \& Grounding}} \\
\cmidrule(lr){2-3} \cmidrule(lr){4-6} \cmidrule(lr){7-8}
& AUROC $\uparrow$ & Macro-F1 $\uparrow$ & ROUGE-L $\uparrow$ & CIDEr $\uparrow$ & ClinEnt-F1 $\uparrow$ & Halluc. $\downarrow$ & mIoU $\uparrow$ \\
\midrule
\multicolumn{8}{l}{\textit{Risk-Only Baselines}} \\
\midrule
XGBoost \cite{chen2016xgboost} & 0.785 & 0.742 & -- & -- & -- & -- & -- \\
TabNet \cite{arik2021tabnet} & 0.791 & 0.750 & -- & -- & -- & -- & -- \\
Late-Fusion (CT+Echo+Tab) & 0.832 & 0.796 & -- & -- & -- & -- & -- \\
\midrule
\multicolumn{8}{l}{\textit{Report Generation Baselines}} \\
\midrule
MCLN-Transformer \cite{chen2020r2gen} & -- & -- & 0.316 & 0.662 & 0.432 & 37.9\% & 0.175 \\
R2Gen-Mamba \cite{sun2024r2genmambaselectivestatespace} & -- & -- & 0.329 & 0.684 & 0.458 & 34.6\% & 0.210 \\
METransformer \cite{2023METransformer} & -- & -- & 0.341 & 0.703 & 0.470 & 33.8\% & 0.234 \\
\midrule
\multicolumn{8}{l}{\textit{Open-Source Multimodal Models}} \\
\midrule
LLaVA-Med-1.5 \cite{li2023llavamed} & 0.804 & 0.773 & 0.388 & 0.821 & 0.614 & 24.2\% & 0.454 \\
Qwen3-VL-8B \cite{yang2025qwen3technicalreport} & 0.849 & 0.823 & 0.401 & 0.868 & 0.695 & 17.9\% & 0.501 \\
InternVL3.5-8B \cite{wang2025internvl35advancingopensourcemultimodal} & 0.861 & 0.836 & 0.409 & 0.889 & 0.724 & 14.7\% & 0.516 \\
\midrule
\multicolumn{8}{l}{\textit{Closed-Source Frontier Models}} \\
\midrule
GPT-4o & 0.868 & 0.844 & 0.412 & 0.901 & 0.756 & 13.8\% & -- \\
Claude-4 Opus & 0.875 & 0.852 & 0.415 & 0.908 & 0.764 & 12.9\% & --\\
Gemini-3 Pro  & 0.883 & 0.860 & 0.418 & 0.914 & 0.772 & 11.5\% & -- \\
\midrule
\multicolumn{8}{l}{\textit{TAVR-VLM (Proposed Framework)}} \\
\midrule
\textbf{R-CGA + Qwen3-VL-8B} & \underline{0.892} & \underline{0.869} & \underline{0.422} & \underline{0.928} & \underline{0.789} & \underline{8.7\%} & \underline{0.603} \\
\textbf{R-CGA + InternVL3.5-8B} & \textbf{0.896} & \textbf{0.874} & \textbf{0.426} & \textbf{0.936} & \textbf{0.801} & \textbf{8.1\%} & \textbf{0.624} \\
\bottomrule
\end{tabular}}
\end{table*}

\subsection{Experimental Setup and Evaluation Metrics}
\label{sec:exp_setup}
To ensure a comprehensive evaluation, we benchmark our framework against four distinct paradigms: risk-only classifiers, medical report generators, open-source MLLMs, and closed-source frontier models, extracting standardized 2D slices for 2D-centric baselines to maintain domain fairness. Our architecture, driven by a 3D ViT for CT extraction and a temporal 2D-CNN for echocardiography, is optimized end-to-end via AdamW across four NVIDIA A100 GPUs with the causal consistency coefficient $\lambda_{\text{causal}}$ empirically set to 0.5. We evaluate the joint modeling capabilities across three critical dimensions: clinical prediction performance (AUROC and Macro-F1), generation fidelity (ROUGE-L \cite{lin2004rouge} and CIDEr \cite{vedantam2015cider}). To rigorously penalize diagnostic fabrication, we define Hallucination Rate (HR) as the percentage of generated clinical entities lacking corresponding multimodal evidence, while spatial anchoring precision is quantified via mean Intersection over Union (mIoU) between the constrained token-level mask $\widetilde{M}_t$ and expert-annotated ROIs.

\subsection{Main Results}
\label{sec:main_results}

Table~\ref{tab:main} presents a comprehensive quantitative comparison on the \textbf{M$^3$TAVR} test set. By benchmarking against four distinct paradigms, we demonstrate that our proposed R-CGA framework establishes a new state-of-the-art across all evaluated clinical and generation dimensions.

\paragraph{Superiority in Clinical Prediction and Generation.}
While current open-source models and cutting-edge closed-source VLMs exhibit formidable general vision-language capabilities, they lack the explicit domain constraints required for high-stakes surgical planning. By integrating our plug-and-play R-CGA module into an 8B-parameter backbone (InternVL3.5-8B), we observe a transformative performance leap. Specifically, our model achieves an AUROC of 0.896 in risk prediction, outperforming the top-performing frontier model, Gemini-3 Pro (0.883). In report generation, our framework surpasses Gemini-3 Pro in CIDEr (0.936 vs.\ 0.914), indicating a higher consensus with clinical experts in utilizing precise TAVR-specific terminology rather than generic descriptions.

\paragraph{Breakthrough in Safety and Spatial Grounding.}
Diagnostic hallucination remains the most critical bottleneck for MLLM adoption in structural heart interventions. As shown in Table~\ref{tab:main}, traditional report generation baselines suffer from severe hallucinations ($>33\%$) and diffuse visual attention (mIoU $< 0.25$). Even the most advanced frontier model, Gemini-3 Pro, yields a hallucination rate of 11.5\%, as its autoregressive generation relies heavily on generic memory priors rather than rigorous visual evidence. 

In stark contrast, the top-down causal bottleneck in R-CGA reduces hallucinations through risk-conditioned grounding. Enforced by this structural prior, our hallucination rate decreases to 8.1\%, while token-to-region grounding precision (mIoU) surges to 0.624. This empirically validates that our model does not merely memorize linguistic distributions; it accurately localizes pathological evidence (e.g., specific calcified leaflets) before emitting the corresponding clinical entities.

\begin{figure*}[t]
    \centering
    \begin{minipage}[b]{0.4\textwidth}
        \centering
        \includegraphics[width=\linewidth]{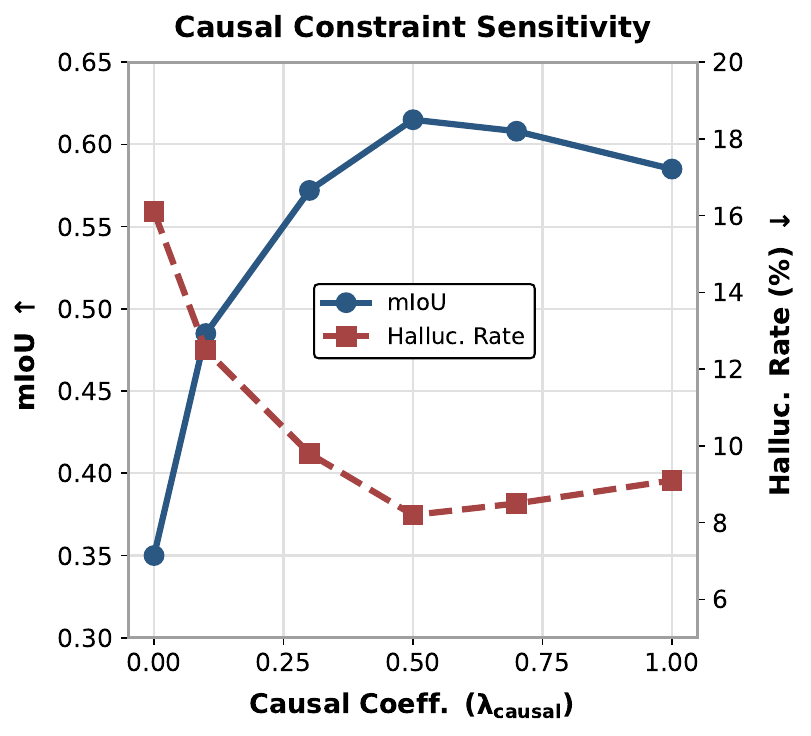}
        \centerline{\small (a) Parameter Sensitivity}
    \end{minipage}
    \hfill
    \begin{minipage}[b]{0.58\textwidth}
        \centering
        \includegraphics[width=\linewidth]{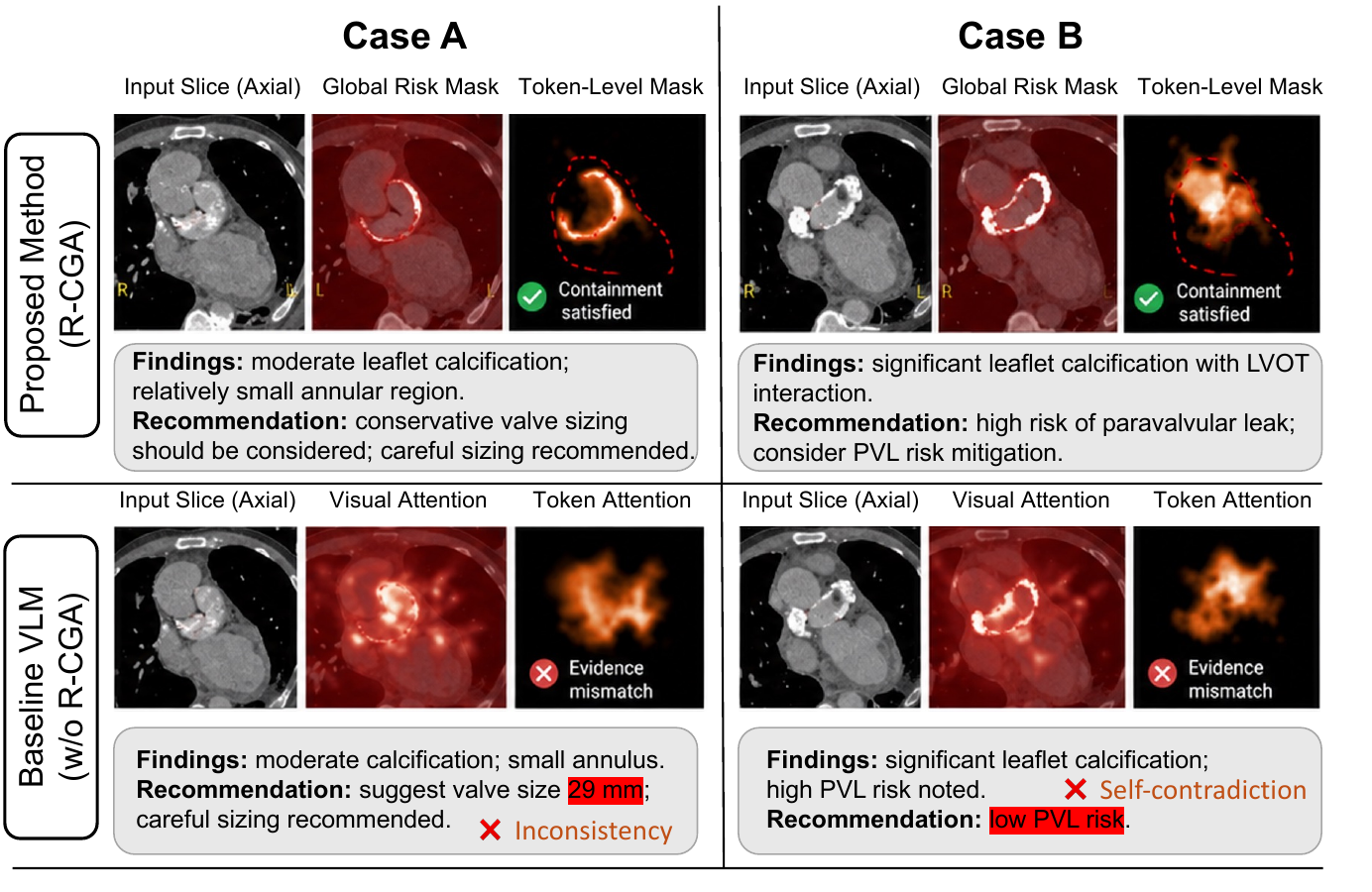}
        \centerline{\small (b) Qualitative Case Study}
    \end{minipage}
    \vspace{-1em}
    \caption{\textbf{Comprehensive Model Analysis.}  \textbf{(a)} Sensitivity of the causal consistency
coefficient $\lambda_{\mathrm{causal}}$ on spatial grounding accuracy and hallucination rate. \textbf{(b)} Qualitative comparison between R-CGA and a baseline VLM
without risk-conditioned grounding.}
    \label{fig:analysis}
    \vspace{-1.5em}
\end{figure*}

\subsection{Ablation Study}
\label{sec:ablation}

\subsubsection{Module Ablation.}
To disentangle the contributions of the R-CGA components, we conduct an ablation study (Table~\ref{tab:ablation_module}). Removing the learnable risk prototypes ($W$) degrades the macro bottleneck representation, lowering AUROC to 0.875. Crucially, bypassing the risk-driven visual purification ($\text{w/o } V_{\text{act}}$) exposes the language model to unconstrained tokens, triggering an immediate spike in hallucination rate (from 8.1\% to 18.4\%) as the generator exploits spurious correlations. Disabling the causal consistency loss ($\lambda_{\text{causal}}=0$) severs the top-down alignment, causing a steep drop in spatial grounding precision (mIoU falls to 0.350). Finally, replacing the stop-gradient mechanism with standard backpropagation ($\text{w/o StopGrad}$) results in degenerate optimization; the model artificially inflates $M_{\text{global}}$ to trivially minimize KL divergence, validating that risk must serve as a strict, unidirectional bottleneck.

\subsubsection{Parameter Sensitivity.}
We analyze the sensitivity of the causal consistency coefficient $\lambda_{\text{causal}}$ (Fig.~\ref{fig:analysis}a). A clear trade-off emerges: increasing $\lambda_{\text{causal}}$ from 0 to 0.5 drastically improves mIoU (from 0.350 to 0.624) while curbing the diagnostic hallucination rate down to 8.2\%. However, overly aggressive penalization ($\lambda_{\text{causal}} > 0.5$) marginally degrades performance, indicating that the language generator becomes too rigidly constrained by the global mask, sacrificing linguistic flexibility. Based on these dynamics, we adopt $\lambda_{\text{causal}} = 0.5$ and a prototype dimension of $d = 512$ for all experiments.

\subsubsection{Qualitative Case Study.}
To intuitively understand how R-CGA suppresses hallucinations, we visualize representative cases comparing R-CGA with a baseline VLM without risk-conditioned grounding in Fig.~\ref{fig:analysis}b. The baseline model often covers diffuse or clinically irrelevant regions, leading to a mismatch of evidence and unsupported recommendations. In contrast, R-CGA first selects a global risk-relevant region and then constrains token-level grounding to this region through support projection. This mechanism reduces unsupported clinical entities and improves the anatomical consistency between the visual evidence and the generated report statements.
\begin{table}[t]
\centering
\caption{Ablation of key R-CGA components on \textbf{M$^3$TAVR}.}
\label{tab:ablation_module}
\resizebox{\linewidth}{!}{
\begin{tabular}{l|cc|cc}
\toprule
\textbf{Variant} & \textbf{AUROC} $\uparrow$ & \textbf{CIDEr} $\uparrow$ & \textbf{Halluc. Rate} $\downarrow$ & \textbf{mIoU} $\uparrow$ \\
\midrule
\textbf{Full TAVR-VLM} & \textbf{0.896} & \textbf{0.936} & \textbf{8.1\%} & \textbf{0.624} \\
\midrule
w/o Risk Prototypes $W$ & 0.875 & 0.890 & 12.5\% & 0.542 \\
w/o Purification ($V_{\text{act}} = V$) & 0.881 & 0.865 & 18.4\% & 0.485 \\
w/o Causal Loss ($\lambda_{\text{causal}}=0$) & 0.885 & 0.872 & 16.1\% & 0.350 \\
w/o Stop-Gradient & 0.888 & 0.895 & 14.3\% & 0.412 \\
\bottomrule
\end{tabular}}
\vspace{-1.5em}
\end{table}

\section{Conclusion}
\label{sec:conclusion}
We presented TAVR-VLM, a pioneering risk-conditioned multimodal framework for hallucination-resistant TAVR report generation. The proposed R-CGA module constructs a model-internal ``Risk $\rightarrow$ Region $\rightarrow$ Word'' structural pathway by using a causal risk bottleneck to guide anatomical evidence selection and token-level grounding. Unlike unconstrained generation, R-CGA projects clinically significant token attention into a risk-defined support mask and penalizes grounding outside this support, thus mitigating evidence-mismatched clinical statements. Experiments on the $\text{M}^3\text{TAVR}$ dataset demonstrate that TAVR-VLM outperforms both specialized medical models and cutting-edge frontier VLMs in risk prediction, report generation, hallucination reduction, and
spatial grounding. Future work will focus on external validation, more detailed clinical endpoint modeling, and broader evaluation across structural heart interventions.

\bibliographystyle{splncs04}
\bibliography{bibliography}
%




\end{document}